\pdfoutput=1

\documentclass[11pt]{article}
\usepackage{float}
\usepackage{makecell}
\restylefloat{table}
\usepackage{booktabs}

\usepackage{ACL2023}

\usepackage{times}
\usepackage{latexsym}
\usepackage{amsmath}
\usepackage{amssymb}
\usepackage{graphicx}

\usepackage[T1]{fontenc}

\usepackage[utf8]{inputenc}

\usepackage{microtype}

\usepackage{inconsolata}

\title{Improving Text Embeddings for Smaller Language Models \\ Using Contrastive Fine-tuning}

\author{Trapoom Ukarapol, Zhicheng Lee, Amy Xin \\
        Department of Computer Science and Technology, Tsinghua University \\
        \texttt{\{ukarapolt10, lizhiche23, xin-x23\}@mails.tsinghua.edu.cn}}

\begin{document}
\maketitle
\begin{abstract}

While Large Language Models show remarkable performance in natural language understanding, their resource-intensive nature makes them less accessible. In contrast, smaller language models such as MiniCPM offer more sustainable scalability, but often underperform without specialized optimization. In this paper, we explore the enhancement of smaller language models through the improvement of their text embeddings. We select three language models, MiniCPM, Phi-2, and Gemma, to conduct contrastive fine-tuning on the NLI dataset. Our results demonstrate that this fine-tuning method enhances the quality of text embeddings for all three models across various benchmarks, with MiniCPM showing the most significant improvements of an average 56.33\% performance gain. The contrastive fine-tuning code is publicly available at \url{https://github.com/trapoom555/Language-Model-STS-CFT}.

\end{abstract}

\section{Introduction}

Text embeddings are vector representations of text data that encodes 
semantic information, enabling machines to understand and process natural language. These embeddings are crucial for a variety of tasks including document classification, 
semantic similarity matching, 
and information retrieval. 
With the recent advancements of Large Language Models (LLMs),
LLMs such as GPT-4 \cite{achiam2023gpt}, LLaMA \cite{touvron2023llama}, and Mistral \cite{jiang2023mistral} have demonstrated robust natural language understanding capabilities due to their large-scale training. Conversely, smaller models like Gemma \cite{team2024gemma}, Phi \cite{li2023textbooks}, and MiniCPM \cite{hu2024minicpm}, while less resource-intensive, often lag in performance without specific optimizations.

This project is motivated by the lack of research focused on enhancing the text embedding capabilities of relatively smaller models. Notably, MiniCPM, a smaller-scale language model, underperforms in generating effective text embeddings without further fine-tuning. Addressing this gap, we aim to conduct experiments to improve the quality of text embeddings generated by MiniCPM, thereby making it a viable option for resource-constrained applications. This project could pave the way for using small scale models on the text embedding task. Furthermore, we compare MiniCPM's performance among other small language models including Gemma and Phi-2 after fine-tuning with the same dataset. 

\section{Related Works}

\paragraph{Text Embeddings} Text embeddings are vector representations of text in a low-dimensional space. These embeddings are designed to encapsulate the semantic meaning of the text, facilitating various downstream tasks including information retrieval, document classification, and similarity matching. Traditional models such as SBERT \cite{reimers2019sentence} and Sentence T5 \cite{ni2021sentence} have aimed to provide a universal framework for encoding the semantic significance of text across different tasks and domains. Moreover, methods such as Contriever \cite{izacard2021unsupervised}, E5 \cite{wang2022text}, and SGPT \cite{muennighoff2022sgpt} have implemented a multi-stage training strategy to enhance the effectiveness of text embeddings. To assess the performance of text embeddings within the proposed model, the benchmark MTEB \cite{muennighoff2022mteb} is utilized. This benchmark includes novel tasks to evaluate the text embedding quality of the models and their abilities to generalize across different domains. 


\paragraph{Contrastive Representation Learning} Contrastive Representation Learning aims to learn effective representation by pulling semantically close neighbors
together and pushing apart non-neighbors \cite{hadsell2006dimensionality}. Early works learn such representation by using contrastive loss \cite{chopra2005learning} and triplet loss \cite{schroff2015facenet}, where only one positive and one negative sample are involved. On top of that, multi-class N-pair loss \cite{sohn2016improved} and InfoNCE loss \cite{oord2018representation} include multiple positive and negative pairs in a batch to improve the semantic representation of the embeddings.

\paragraph{Lightweight LLMs} Despite the remarkable performance of LLMs in text generation tasks, the cost of training, serving and maintaining such large scale model is huge. Therefore, several lightweight LLMs within the range of 1 to 3 billion parameters have been proposed to tackle this issue.  Notable examples include the Phi series \cite{li2023textbooks}, Gemma \cite{team2024gemma} and MiniCPM \cite{hu2024minicpm}. These lightweight language models require fewer resources, thereby reducing the barriers to entry for conducting studies with smaller-scale models.

\paragraph{LLM Fine-tuning}

In the recent year, Pre-training and Fine-tuning Paradigm has been successfully proven to be better than training from scratch. The foundational work, Bidirectional Encoder Representations from Transformers (BERT) \cite{devlin2019bert} established the pre-training and fine-tuning paradigm, where models are first pre-trained on a large corpus of text and then fine-tuned on specific downstream tasks. This approach has become a cornerstone for many subsequent models, including GPT-3 \cite{brown2020language}, which demonstrated that increasing model size and the amount of pre-training data can lead to significant improvements in performance across a wide range of tasks. Recent research has focused on making fine-tuning more efficient, particularly for large-scale models. Techniques such as Adapter modules, proposed by \cite{houlsby2019parameterefficient}, involve inserting small trainable layers within the pre-trained model, allowing for task-specific adaptation without modifying the entire model's parameters. This approach significantly reduces the computational cost of fine-tuning.
Another notable technique is Low-Rank Adaptation (LoRA), introduced by \cite{hu2021lora}, which adapts the weights of the model using low-rank matrix factorization. This method achieves substantial parameter efficiency, enabling the fine-tuning of large models with a fraction of the computational resources typically required.


\section{Methodology}

Our approach aims to address the problem of Semantic Textual Similarity (STS) in the English language by employing smaller language models, thereby ensuring that our solution is both efficient and scalable. 
To enhance the text embedding capabilities of these smaller models, we utilize a contrastive fine-tuning approach. Contrastive fine-tuning involves training the model to distinguish between similar and dissimilar pairs of text. This method helps the model to generate more accurate and contextually relevant embeddings, which are essential for assessing semantic similarity. The details of contrastive fine-tuning are described in section \ref{sec:cft}.

Additionally, we adopt a parameter-efficient fine-tuning technique. This technique is designed to achieve optimal performance while minimizing the computational resources required. By using the low-rank adaptation (LoRA) \cite{hu2021lora} method, we ensure that our approach remains computationally feasible even with limited hardware resources. This is particularly important for fine-tuning the models in the scenario where computational efficiency is a key concern.

\subsection{Dataset} 
The processed NLI dataset \cite{oord2018representation} is adopted as our training dataset. There are approximately 275k samples in the dataset. Each entry includes a premise $x_i$, along with its corresponding entailment $x_i^+$ and contradiction $x_i^-$, thus forming a triplet $(x_i, x_i^+, x_i^-)$

\subsection{Language Model Choices}

We conduct our experiments on language models with fewer parameters up to 2B, including Gemma \cite{team2024gemma}, Phi-2 \cite{li2023textbooks}, and MiniCPM \cite{hu2024minicpm} to explore their capabilities on the text embedding task.

\subsection{Contrastive Fine-tuning} 
\label{sec:cft}

In this study, we leverage the language understanding capabilities of language models (LMs) and enhance their text embedding quality using a contrastive fine-tuning paradigm. To achieve higher text embedding quality with limited computational resources, we adopt a parameter-efficient fine-tuning technique, LoRA \cite{hu2021lora}, as our fine-tuning method. The details of how text embeddings are extracted from the  Language Model (LM) and the training objective are described in the following sections.

\paragraph{Embedding Vector Extraction}
Given a pretrained language model and a prompt, we augment the prompt by appending the <EOS> token. Subsequently, we input this modified prompt into the language model. The embedding vector can then be acquired by extracting the vector corresponding to the final <EOS> token from the last layer.

\paragraph{Training Objective}
We utilize the standard InfoNCE objective \cite{oord2018representation} with in-batch negatives and hard negatives. The training objective can be described by the following expression.

\begin{equation} \label{eq:infonce_loss}
    \min  - \log \frac{e^{\text{sim}(\textbf{h}_i, \textbf{h}_i^+) / \tau}}{\sum_{j=1}^N \left( e^{\text{sim}(\textbf{h}_i, \textbf{h}_j^+) / \tau }+ e^{\text{sim}(\textbf{h}_i, \textbf{h}_j^-) / \tau} \right)}
\end{equation}
where $\textbf{h}_i$ denotes an embedding vector of a premise $x_i$, $\tau$ denotes a temperature and $\text{sim}(\textbf{h}_i, \textbf{h}_i^+)$ computes the cosine similarity between embedding vectors $\textbf{h}_i$ and $\textbf{h}_i^+$.
 

\section{Experiments}
Given a pair of sentences, the STS task aims to measure the similarity score of the embeddings embeded by the models where higher scores signify greater similarity. Specifically, we measure the distance between embeddings using cosine similarity, then calculate the correlation with ground truth similarities using Spearman correlations. 


We compare the performance of MiniCPM and baseline models such as Gemma and Phi-2 on 9 benchmarks, namely STS12 \cite{agirre2012semeval}, STS13 \cite{agirre2013sem}, STS14\footnote{\url{https://alt.qcri.org/semeval2014/tas
k10/}}, STS15\footnote{\url{https://alt.qcri.org/semeval2015/tas
k2/}}, STS16\footnote{\url{https://alt.qcri.org/semeval2016/tas
k1/}}, STS17\footnote{\url{https://alt.qcri.org/semeval2017/tas
k1/}}, STSBenchmark, BIOSSES\footnote{\url{https://tabilab.cmpe.boun.edu.tr/BIO
SSES/DataSet.html}} and SICK-R \cite{agirre2014semeval}. These datasets consist of pairs of sentences with similarity scores from 0 to 5. To ensure fairness, both MiniCPM and the baseline models are fine-tuned on the same processed NLI dataset.

\paragraph{STS12, STS13, STS14, STS15, STS16, STS17, STSBenchmark} This set of STS  benchmarks incorporate sentences derived from various sources including image captions, news headlines, and user forums. These datasets range from 1,000 to 20,000 sentences. 


\paragraph{BIOSSES} The BIOSSES benchmark contains 100 sentence pairs from the biomedical field.

\paragraph{SICK-R} The Sentences Involving Compositional Knowledge (SICK) dataset comprises a substantial corpus of 100,000 sentence pairs, each characterized by its lexical, syntactic, and semantic richness.

\begin{table*}[ht]
\centering
\begin{tabular}{ll}
\toprule
\textbf{Fine-tuning Details} & \textbf{Value} \\
\midrule
Loss & InfoNCE \\
Batch Size & 60 \\
Learning Rate & 5e-05 \\
LoRA Rank & 8 \\
LoRA Alpha & 32 \\
LoRA Dropout & 0.1 \\
Mixed Precision & bf16 \\
Max Epoch & 1 \\
Learning Rate Scheduler & CosineAnnealingLR \\
Learning Rate Warmup Steps & 100 \\
GPU & RTX3090 \\
Num GPUs & 4 \\
\bottomrule
\end{tabular}
\caption{Contrastive fine-tuning configuration details.}
\label{tab:fine_tuning_details}
\end{table*}

\subsection{Main Experiment Results}
Table \ref{tab:model_performance} presents the evaluation results of three models: Gemma, Phi-2, and MiniCPM across all 9 benchmarks, using Spearman correlations of cosine similarities post-supervised fine-tuning (SFT). The results demonstrate that MiniCPM consistently outperforms the other models on all benchmarks.

Specifically, MiniCPM achieves the highest Spearman correlations across all datasets, including notable results on STS12 (76.38\%), STS13 (87.61\%), and STS17 (89.96\%). Gemma showed competitive performance but trailed slightly behind MiniCPM in every benchmark, with its highest correlation observed on STS17 (88.22\%). Phi-2, on the other hand, exhibited the lowest performance among the three models, with its best result on STS17 (80.20\%).

The overall performance metrics further illustrate this trend, with MiniCPM achieving an average Spearman correlation of $83.84\% \pm 4.27$, which is superior to Gemma's $81.56\% \pm 4.83$ and Phi-2's $70.89\% \pm 6.91$. These results indicate that while all models benefited from the fine-tuning process, MiniCPM's architecture and training regimen enabled it to achieve superior alignment with human judgment across diverse datasets.

In conclusion, the evaluation underscores MiniCPM's robustness and effectiveness in capturing semantic similarities, making it a reliable choice for tasks requiring nuanced understanding of textual data.

The fine-tuned versions of these three models have been published and are available for access here\footnote{\url{https://huggingface.co/collections/trapoom555/small-lms-text-embedding-663b3ec87527788a577f6852}}.

\begin{table*}[ht]
\centering
\begin{tabular}{lccc}
\toprule
\textbf{Benchmark} & \textbf{Gemma} & \textbf{Phi-2} & \textbf{MiniCPM} \\
\midrule
STS12 & 75.80 & 61.62 & \textbf{76.38} \\
STS13 & 85.45 & 71.87 & \textbf{87.61} \\
STS14 & 80.08 & 60.46 & \textbf{81.55} \\
STS15 & 85.02 & 71.18 & \textbf{87.33} \\
STS16 & 83.33 & 74.77 & \textbf{85.25} \\
STS17 & 88.22 & 80.20 & \textbf{89.96} \\
STSBenchmark & 85.61 & 79.46 & \textbf{86.51} \\
BIOSSES & 73.83 & 64.06 & \textbf{80.05} \\
SICK-R & 76.69 & 74.37 & \textbf{79.87} \\
\bottomrule
Overall & $81.56\pm4.83$ & $70.89\pm6.91$ & \textbf{83.84}$\pm$\textbf{4.27} \\
\hline
\end{tabular}
\caption{Spearman correlations of cosine similarities of various models after SFT across different datasets.}
\label{tab:model_performance}
\end{table*}

\subsection{Ablation Studies}

\subsubsection{Model performance before SFT}
In this study, we aim to compare the performance of Gemma, Phi-2 and MiniCPM
before and after applying SFT (fine-tuning). The results are detailed in Tables \ref{tab:model_performance} and \ref{tab:model_performance_before_sft}.

Table \ref{tab:model_performance_before_sft} presents the spearman correlations of cosine similarities for the models prior to SFT across various benchmarks. Gemma consistently outperforms Phi-2 and MiniCPM on all benchmarks, achieving the highest overall correlation of 
$56.63\% \pm 7.89$. Phi-2 and MiniCPM follow with overall correlations of $32.21\% \pm 11.40$ and $27.51\% \pm12.76$, respectively.

Comparing the results before and after SFT, we observe significant performance improvements on all three models. Notably, MiniCPM demonstrates the most substantial improvement, increasing its overall correlation by approximately 56 points. This indicates that SFT has a pronounced positive impact on MiniCPM’s performance.

In summary, while all models benefit from SFT, 
MiniCPM stands out as the top performer after SFT.

\begin{table*}
\centering
\begin{tabular}{lccc}
\toprule
\textbf{Benchmark} & \textbf{Gemma} & \textbf{Phi-2} & \textbf{MiniCPM} \\
\midrule
STS12 & \textbf{43.83} & 23.04 & 7.27 \\
STS13 & \textbf{66.36} & 20.79 & 18.38 \\
STS14 & \textbf{49.57} & 17.06 & 15.04 \\
STS15 & \textbf{57.40} & 24.56 & 32.24 \\
STS16 & \textbf{70.13} & 48.68 & 39.79 \\
STS17 & \textbf{58.34} & 41.43 & 33.63 \\
STSBenchmark & \textbf{57.36} & 37.87 & 33.91 \\
BIOSSES & \textbf{48.67} & 28.04 & 18.03 \\
SICK-R & \textbf{58.02} & 48.40 & 49.30 \\
\midrule
Overall & \textbf{56.63}$\pm$\textbf{7.89} & $32.21\pm11.40$ & $27.51\pm12.76$ \\
\bottomrule
\end{tabular}
\caption{Spearman correlations of cosine similarities of various models before SFT across different datasets.}
\label{tab:model_performance_before_sft}
\end{table*}

\subsubsection{Impact of learning rate}
In this study, we aim to study the impact of learning rate during training. Results from Table~\ref{tab:learning_rate} shows that learning rate 5e-5 achieves the best result among the others, whereas learning rate 5e-3 achieves the lowest spearman correlations score, indicating the model suffers from training instability and is underfitting.

\begin{table}[htbp]
\centering
\begin{tabular}{lcc}
\toprule
\textbf{Learning Rate} & \textbf{MiniCPM} \\
\midrule
5e-3 & $36.19 \pm 8.91$  \\
5e-4 & $82.41 \pm 5.01$  \\
5e-5 & $82.31 \pm 5.27$ \\
\bottomrule
\end{tabular}
\caption{Experimental results of varying learning rate during training. The values shown are the average spearman correlations of cosine similarities across all 9 test datasets}
\label{tab:learning_rate}
\end{table}

\subsubsection{How does prompting affect model performance?}

In our main experiments shown in Tables \ref{tab:model_performance} and \ref{tab:model_performance_before_sft}, we tested MiniCPM's embeddings without any additional prompts. 
Yet, it is known that employing prompting techniques may further leverage a language model's performance.
Hence, we conducted an ablation study on MiniCPM with prompt-tuning.

We adopted the Explicit One-Word Limitation prompting technique \cite{jiang2023scalingsentenceembeddingslarge}, along with two other prompts:

\begin{enumerate}
    \item "This sentence: \{original\_sentence\} means in one word: "
    \item "This sentence \{original\_sentence\} means: "
    \item "\{original\_sentence\} is: "
\end{enumerate}

During testing, \{original\_sentence\} is replaced by the sentence from the current test case.

We tested all 3 prompts on both our fine-tuned and original MiniCPM models. The results are shown in Table \ref{tab:prompting}.

\begin{table}[htbp]
\centering
\begin{tabular}{lcc}
\toprule
\textbf{Prompt} & \textbf{MiniCPM (tuned)} & \textbf{MiniCPM} \\
\midrule
No Prompt & \underline{83.84} & 27.51 \\
\midrule
Prompt 1 & 81.81 & \textbf{34.76} \\
Prompt 2 & 83.03 & 26.92 \\
Prompt 3 & \textbf{84.32} & \underline{28.79} \\
\bottomrule
\end{tabular}
\caption{Results of our prompt-tuning experiments. The values shown are the average spearman correlations of cosine similarities across all 9 test datasets under the specified prompt.}
\label{tab:prompting}
\end{table}

As the results show, the original MiniCPM model performs best with Prompt 1, reaching a +7.25\% gain in average performance. The original model experiences performance enhancements with both Prompt 1 and Prompt 3. 
In contrast, our fine-tuned MiniCPM model only achieves a minor +0.48\% performance gain with Prompt 3, and is negatively impacted by both Prompt 1 and Prompt 2. 
Such results suggest that prompt-tuning is effective in certain conditions but harmful in others, and it is more effective to the original MiniCPM model than to our fine-tuned MiniCPM model. 
We hypothesize that this may be due to our fine-tuned model's preference towards the distribution of the original sentences instead of sentences augmented with prompts, since the model has been exposed to more sentences of the original format during fine-tuning.

\subsubsection{Training data efficiency}
\label{sec:train_data_eff}

\begin{figure}[htbp]
  \centering
  \includegraphics[width=1\linewidth]{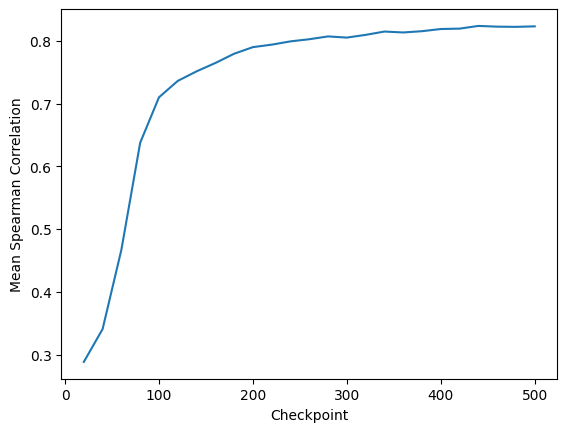}
  \caption{Average performance of model checkpoints across all 9 test sets. The values are the average spearman correlations of cosine similarities across all 9 test sets. The model converged after checkpoint 200.}
  \label{fig:data_effi}
\end{figure}

In this study, we aim to study training data efficiency by measuring how much training data is required to achieve convergences. To do so, we measure evaluate the every 20 checkpoints until the Spearman correlation coverged since lesser training step equivalent to lesser training data used. The results are recorded in Figure~\ref{fig:data_effi}. 

Our results show that after 200 training steps, MiniCPM has already earned a average performance gain of +50\% over our 9 test sets, indicating that it underwent the majority of its capability growth during the first 200 steps.
Furthermore, the model begins to converge at around 200 steps, reaching an average performance of 82.31\% at Checkpoint-500. 
This shows that MiniCPM demonstrates a relatively high training efficiency with great convergence speed.

\subsubsection{Impact of hard negatives penalty}
In the context of contrastive learning, the selection of hard negatives plays a critical role in enhancing the effectiveness of the training process. To study the impact of the hard negatives on the performance of MiniCPM after SFT, we remove the penalization on the hard negative samples by modifying the objective function in Equation \ref{eq:infonce_loss} to become:

\begin{equation} \label{eq:infonce_loss_noneg}
    \min  - \log \frac{e^{\text{sim}(\textbf{h}_i, \textbf{h}_i^+) / \tau}}{\sum_{j=1}^N \left( e^{\text{sim}(\textbf{h}_i, \textbf{h}_j^+) / \tau }\right)}
\end{equation}

Table \ref{tab:minicpm_performance_no_neg} compares the Spearman correlations of cosine similarities of the MiniCPM model with and without penalization on the hard negative samples. The results indicate that incorporating penalization generally improves performance across most benchmarks. Specifically, the MiniCPM model shows an increase in performance on all benchmarks except STS17 when hard negative samples are penalized.

For instance, the Spearman correlation for STS13 improves from 85.98\% to 87.61\%, which is a notable increase. Similarly, STS15 sees a rise from 85.66\% to 87.33\%, and STS14 improves from 78.52\% to 81.55\%. These results suggest that penalizing hard negative samples helps the model better distinguish between similar and dissimilar pairs, leading to higher correlation scores.

However, there is a slight decrease in performance for STS17, where the correlation drops from 90.23\% to 89.96\%. This anomaly might indicate that for some datasets, the penalization might not always lead to improvements, potentially due to the nature of the data or the specific characteristics of the benchmark.

Overall, the average performance across all benchmarks increases from 81.97\% to 83.84\%. This demonstrates that penalizing hard negative samples not only boosts the average performance but also leads to more consistent results, as evidenced by the lower standard deviation.

\begin{table}[htbp]
\centering
\begin{tabular}{lcc}
\toprule
\textbf{Benchmark} & \makecell{\textbf{MiniCPM} \\ \textbf{(No Hard Neg)}} & \textbf{MiniCPM} \\
\midrule
STS12 & 75.49 & \textbf{76.38} \\
STS13 & 85.98 & \textbf{87.61} \\
STS14 & 78.52 & \textbf{81.55} \\
STS15 & 85.66 & \textbf{87.33} \\
STS16 & 84.69 & \textbf{85.25} \\
STS17 & \textbf{90.23} & 89.96 \\
STSBenchmark & 85.48 & \textbf{86.51} \\
BIOSSES & 78.31 & \textbf{80.05} \\
SICK-R & 73.35 & \textbf{79.87} \\
\midrule
Overall & $81.97\pm5.37$ & \textbf{83.84}$\pm$ \textbf{4.27} \\
\bottomrule
\end{tabular}
\caption{Spearman correlations of cosine similarities of MiniCPM without penalty on the hard negative samples. The 'MiniCPM (No Hard Neg)' column represents results from training without penalization on hard negatives, whereas the 'MiniCPM' column represents results from training with hard negative penalty.}
\label{tab:minicpm_performance_no_neg}
\end{table}

\section{Conclusion}

In this project, we enhanced the text embedding capabilities of MiniCPM through contrastive fine-tuning using the NLI dataset. 
Our results demonstrate that MiniCPM achieved a significant performance gain of 56.33\% and outperformed 2 other models, Gemma and Phi-2, across 9 STS datasets. 
Additionally, we conducted multiple ablation studies to delve deeper into our method, examining aspects such as prompt-tuning, training efficiency, and the effects of incorporating a hard negatives penalty in the objective function. 
Our research contributes to the enhancement of text embedding qualities in smaller-scale LLMs, making them more robust and reliable for further applications.

\section{Acknowledgement}

We would like to formally acknowledge the Natural Language Processing Spring 2024 course for providing us with the foundational knowledge and the opportunity to undertake this project. We also extend our sincere gratitude to Professor Huaping Liu and the Tsinghua Knowledge Engineering Group for supplying the essential computational resources for our work.

\bibliography{anthology,custom}
\bibliographystyle{acl_natbib}




\appendix

\section{Fine-tuning Details}

In contrastive learning, a large batch size is required to improve training stability. In this work, we aim to enhance text embedding performance with limited computational resources. Therefore, we implement several training techniques to achieve this goal. We employ LoRA \cite{hu2021lora}, a parameter-efficient fine-tuning method, to significantly decrease the GPU VRAM requirements for a single batch. Consequently, more batches can fit into a single GPU. To further reduce GPU VRAM demand, we adopt the bf16 mixed-precision training technique, which also increases training speed while maintaining numerical stability. Additionally, we leverage multi-GPU processing to amplify batch size across GPUs using the distributed data parallelism (DDP) technique. Furthermore, we incorporate a learning rate scheduler with CosineAnnealingLR \cite{cosineAnnealingLR} and learning rate warm-up to optimize the training process.

All fine-tuning details listed in Table \ref{tab:fine_tuning_details}. With these settings, the model can be effectively fine-tuned using limited resources, ensuring efficiency and cost-effectiveness. The entire fine-tuning process involves 13,780 optimization steps to cover the entire dataset.

\begin{figure}[htbp]
  \centering
  \includegraphics[width=0.9\linewidth]{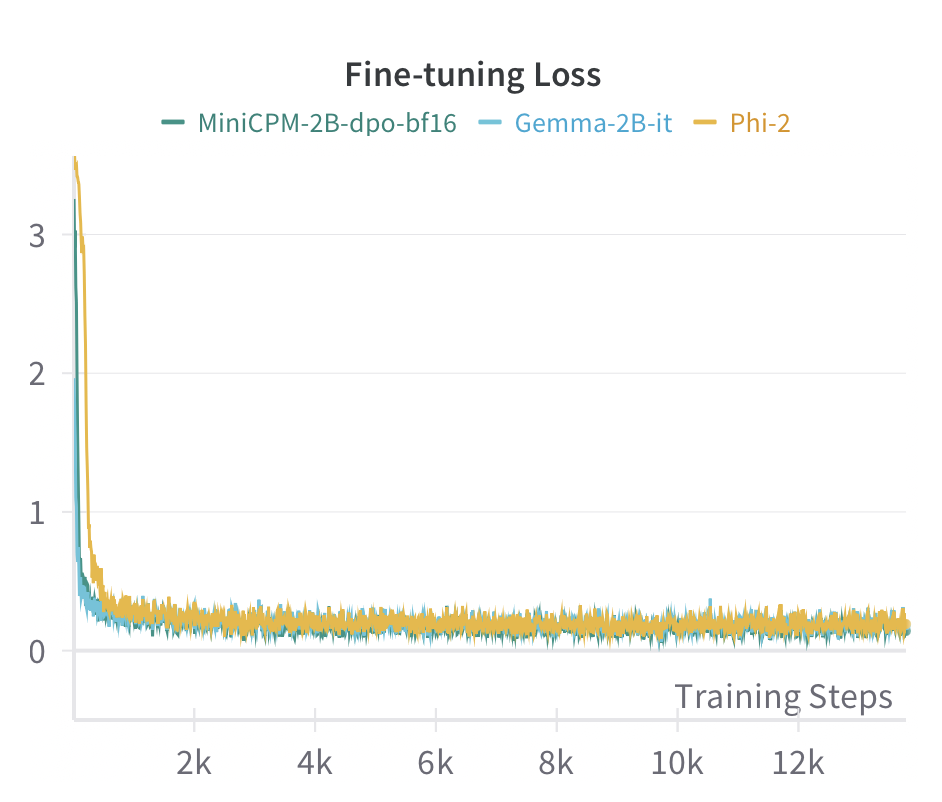}
  \caption{Fine-tuning loss over training steps.}
  \label{fig:loss}
\end{figure}

Figure \ref{fig:loss} illustrates the fine-tuning loss curves of the three selected models: Gemma, Phi-2, and MiniCPM. These curves show that the loss converges very quickly at the beginning of the fine-tuning process, indicating a rapid improvement in model accuracy during the initial stages. We observed this phenomenon and further examined the training data efficiency in the ablation study \ref{sec:train_data_eff}.

\end{document}